\documentclass{article}

\usepackage{PRIMEarxiv}

\usepackage[utf8]{inputenc} 
\usepackage[T1]{fontenc}    
\usepackage{hyperref}       
\usepackage{url}            
\usepackage{booktabs}       
\usepackage{amsfonts}       
\usepackage{nicefrac}       
\usepackage{microtype}      
\usepackage{lipsum}
\usepackage{fancyhdr}       
\usepackage{graphicx}       
\usepackage{makecell}
\usepackage{array}
\usepackage{multirow}
\usepackage{amsmath}
\graphicspath{{media/}}     

\title{Facilitating large language model Russian adaptation 
with Learned Embedding Propagation
\thanks{Preprint version of an article published in the Journal of Language and Education. Copyright held by the owner/author(s). Publication rights licensed to the Journal of Language and Education.} 
}

\author{
  Mikhail Tikhomirov \\
  Lomonosov Moscow State University \\
  Moscow \\
  Russia\\
  \texttt{tikhomirov.mm@gmail.com} \\
   \And
  Daniil Chernyshev \\
  Lomonosov Moscow State University \\
  Moscow \\
  Russia\\
  \texttt{chdanorbis@yandex.ru} \\
}

\begin{document}
\maketitle

\begin{abstract}
\textbf{Background:} Recent advancements in large language model (LLM) technologies have introduced powerful open-source instruction-tuned LLMs that match the text generation quality of leading models like GPT-4. Despite accelerating LLM adoption in sensitive-information environments, the lack of disclosed training data hinders replication and makes these achievements exclusive to specific models.

\textbf{Purpose:} Given the multilingual nature of the latest iteration of open-source LLMs, the benefits of training language-specific LLMs diminish, leaving computational efficiency as the sole guaranteed advantage of this computationally-expensive procedure. This work aims to address the language-adaptation limitations posed by restricted access to high-quality instruction-tuning data, offering a more cost-effective pipeline.

\textbf{Method:} To tackle language-adaptation challenges, we introduce Learned Embedding Propagation (LEP), a novel method with lower training data requirements and minimal disruption of existing LLM knowledge. LEP employs an innovative embedding propagation technique, bypassing the need for instruction-tuning and directly integrating new language knowledge into any instruct-tuned LLM variant. Additionally, we developed Darumeru, a new benchmark for evaluating text generation robustness during training, specifically tailored for Russian adaptation.

\textbf{Results:} We applied the LEP method to adapt LLaMa-3-8B and Mistral-7B for Russian, testing four different vocabulary adaptation scenarios. Evaluation demonstrates that LEP achieves competitive performance levels, comparable to OpenChat 3.5 and LLaMa-3-8B-Instruct. Further improvements were observed through self-calibration and additional instruction-tuning steps, enhancing task-solving capabilities beyond the original models.

\textbf{Conclusion:} LEP offers a viable and efficient alternative to traditional language-specific instruction-tuning, significantly reducing the costs associated with language adaptation while maintaining or surpassing the performance benchmarks set by contemporary LLMs.
\end{abstract}


\section{Introduction}
Emergence of universal instruct-tuned large language models (LLM) such as ChatGPT (Ouyang, 2022) has substantially accelerated the development of natural language processing technologies. However, despite the remarkable achievements in zero-shot task solving, the close-source nature of such models prevented their adoption in the areas with sensitive or exclusive information where any risk of data-leak jeopardizes the integrity of the business process. As a result the rising demand for open-source alternatives drove the researchers to derive methods for knowledge distillation of state-of-the-art LLMs. One of the first approaches was Alpaca (Taori, 2023) which used ChatGPT to synthesize the instruct-tuning data for open-source foundation LLM LLaMA (Touvron, 2023a). While Alpaca was far from state-of-the-art this inspired the creation of more advanced schemes like BactrianX (Li, 2023) that augmented the synthesis process with cross-lingual machine translation which in turn enabled training of open-source multilingual chatbots. However, with release of GPT-4 (Achiam, 2023) which excelled in multilingual setting it became possible to integrate the explicit translation step into instruction synthesis pipeline thus increasing accessibility of knowledge distillation. This has led to creation of series language-specialized instruction-tunes of open-source LLMs such as Saiga (Gusev, 2023), PolyLM (Wei, 2023), Vikhr (Nikolich, 2024), LLAMMAS (Kuulmets, 2024). 

With increasing instruction synthesis quality the open-source language-specific LLMs were closing the gap with the state-of-the-art closed-source solutions eventually hitting the performance ceiling of conventional instruction-tuning (Cui, 2023) due to low utilization of inherent English contextual knowledge which is dominant in state-of-the-art pre-trained open-source LLMs (Touvron, 2023b; Jiang, 2023; Dubey, 2024). As a possible solution researchers (Zhu, 2023; Li, 2024; Chai, 2024) proposed enriching the instruction-tuning datasets with translation tasks which are designed to align new language knowledge with the existing English semantic representations. However, it was shown by Ranaldi (2023) and Husain (2024) that the cause of alignment issue is likely to lie with the inefficiency of tokenization algorithm which can be addressed either by building a new language-specific token vocabulary or by recycling the English tokens for Romanized language representation. 

Inspired by works of Lakew (2018), Kuratov (2019), Rust (2021) \& Yang (2022) on vocabulary adaptation for encoder models Cui et al. (2023) proposed language-specific continued pre-training pipeline for full LLM language adaptation which paired with instruct-tuning on synthesized examples allowed to create Chinese LLaMa, the first open-source model to reach the performance level of ChatGPT with substantially improved computation efficiency thanks to Chinese-adapted tokenization vocabulary. This approach was studied in detail by Tikhomirov (2023) for LLaMa-2 (Touvron, 2023b) adaptation to Russian language and it was shown that semantic alignment efficiency can be further improved with morphologically accurate tokenization algorithm. Moreover, the full LLM language adaptation pipeline was shown by Nguyen (2023) to outperform state-of-the-art closed-source counterparts on low-resource languages due to their bias towards popular languages. 

While the current iteration of language adaptation algorithm is relatively cost-efficient, the benefit of developing language adapted LLMs is falling amid the rapid development of LLM technology and multilingual specialization of open-source options. At the same time it becomes common to release instruction-tuned models (Jiang, 2023; Dubey 2024) that perform on par with closed-source state-of-the-art counterparts without disclosing the instruction-tuning data the quality of which is the major factor of resulting LLM task-solving capabilities (Zhou 2024). Collecting data of such quality requires a considerable investment in human annotation to an extent that only large organizations can afford creation of such datasets (Dubey 2024). If a language specific counterpart of a high quality instruction dataset is unavailable the result of full language adaptation will only have the benefit of higher computational performance as an inferior instruction-tuning data will lead to inferior task-solving performance. 

\begin{figure}
  \centering
  \includegraphics[scale=0.5]{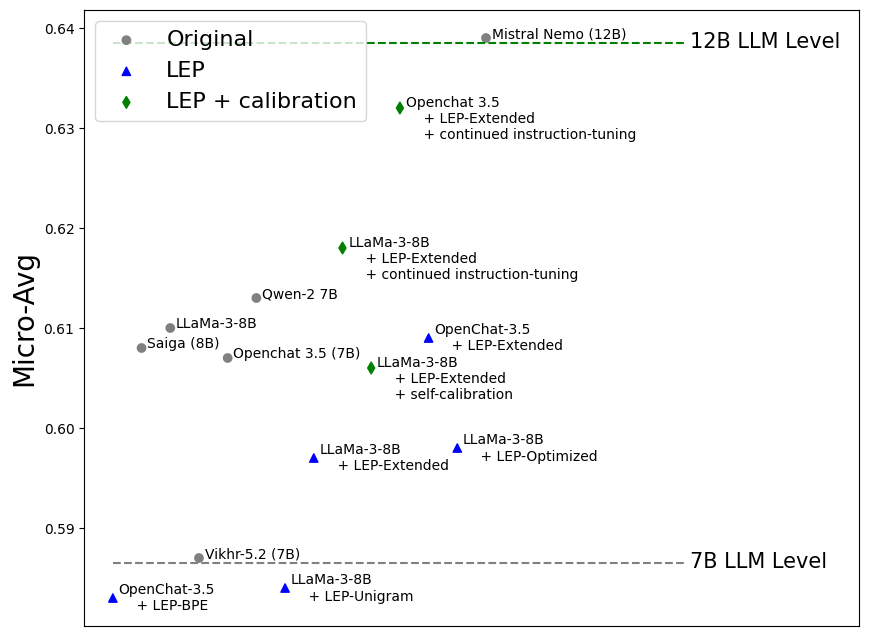}
  \caption{Performance comparison of proposed adaptation method on Darumeru benchmark}
  \label{fig:fig1}
\end{figure}

To cut the language adaptation costs and enable direct language adaptation of instruction-tuned LLM we propose an updated pipeline for language adaptation, Learned Embedding Propagation. Unlike the original full LLM language adaptation pipeline (Cui, 2023), our method requires less data and computational resources due to limited pre-training impact on model parameters which is compensated by novel ad-hoc embedding propagation procedure that allows to skip the instruction-tuning step and instead implant the new language knowledge directly into any existing instruct-tuned variant. To further facilitate the Russian adaptation we developed a new lightweight benchmark for train-time evaluation of LLM text generation robustness, Darumeru. We test Learned Embedding Propagation pipeline on Mistral-7B and LLaMa-3-8B LLMs for 4 Russian tokenization variants. The evaluation results (Figure 1) demonstrate that despite lower parametrization our language-adaptation method manages not only to regain the original quality of the instruction tune but in some cases even outperform it by a significant margin.  Additional case-study experiments on improving the best language-adapted models with continued instruct-tuning and self-calibration also confirm the superiority of our language-adapted models, pushing their performance beyond existing counterparts.

\section{Method}
\subsection{Model Language Adaptation}
Following the previous work on LLM lingual adaptation (Cui, 2023; Tikhomirov, 2023) we first optimize model vocabulary for better alignment with Russian language morphology and then continue the pre-training process on a large corpora of Russian texts of various genres and topics.
Formally the model adaptation consists of 3 steps:
\begin{enumerate}
\item Tokenization training;
\item Model embedding initialization;
\item Continued pre-training of new embeddings (both input and output).
\end{enumerate}

\subsubsection{Tokenization training}
Since there are no best practices for vocabulary optimization we consider 4 options for tokenization training:
\begin{itemize}
\item BPE - fully substituting the tokenization vocabulary by rebuilding the BPE tokenization algorithm (Vries, 2021), which is used in the majority of state-of-the-art LLMs.
\item Unigram - fully substituting the tokenization vocabulary with morphologically accurate tokenization obtained with Unigram algorithm (Tikhomirov, 2023).
\item Extension - extending the original BPE vocabulary by first building a new BPE vocabulary for Russian corpora and then merging it with the original (Cui, 2023).
\item Optimization - refactoring the existing BPE vocabulary by reducing it to the most common 50\% tokens of Russian corpora and then subsequent \item Extension to the original size. (considered only for LLMs with extensive English vocabulary).
\end{itemize}

\subsubsection{Embedding Initialization}
Previous work on LLM language adaptation (Cui, 2023; Tikhomirov, 2023; Nguyen, 2023) found simple averaging of embeddings of overlapping subtokens to be a sufficient solution for embedding initialization. Formally, given embedding vectors of old $v_{old}$ and new $v_{new}$ tokenization vocabularies the new embeddings are initialized as the following:

\begin{equation}
v_{\text{new}}(t^n_i) = \frac{1}{K} \sum_{j=1}^{K} v_{\text{old}}(t^o_j);
\end{equation}
\begin{equation}
tokenize_{old}(t^n_i) = [t^o_1, \dots, t^o_K].
\end{equation}

where $tokenize_{old}$ is the original tokenization function,$t^n_i$ is token in new vocabulary, $t^o_j$ is a token in original vocabulary. 

While there are more advanced initialization techniques, recent studies on design choices for LLM language adaptation (Tejaswi, 2024) concluded that embedding averaging has the best expected adaptation quality and the performance gap with task-tailored methods is within standard deviation of task evaluation protocol. Therefore for all experiments we use the described subtoken averaging embedding initialization strategy.

\subsubsection{Continued pre-training}
The main issue with embedding initialization is that despite introduction of new tokens the LLM retains the habit to use the tokens that were present in the original tokenization. As a result the model computational performance of text generation remains the same as the model tends to use more tokens per word than it is expected while also misinterpreting the new tokens due to homonymy of token context. 

To alleviate the issue the common tactic is to train the newly initialized embeddings on adaptation language corpora using the same pre-training task as LLM, which is causal language modeling. In this task the input text is broken into sequences of tokens of increasing size all of which start from the beginning and the model is asked to predict for each sequence the next possible token. The model optimization is done using simple cross-entropy loss thus any text corpora can be used for the pre-training task. 

Continued pre-training of embeddings only allows the model to tailor those embeddings for inner semantics thus redistributing the existing language knowledge among the newly introduced tokens. However, some researchers (Cui, 2023; Tikhomirov 2024) argued that pre-training embeddings only may be insufficient for proper model-vocabulary alignment and intermediate model layers must be also trained. On the other hand, increasing the number of trained model parameters reduces the training process stability which in turn substantially raises the data size requirements and computational costs of training procedure. As the middle ground we complement embedding pre-training with a post-training layer alignment procedure that recycles existing fine-tunes of the adapted model.

\subsection{Learned Embedding Propagation}
The issue of cost-efficient knowledge transfer for language adapted models has been studied before in the context of encoder models. To solve the absence of task-tuning dataset in the target language Artetxe et al. (2019) proposed a simple algorithm for transferring task-solving knowledge to BERT models: 
\begin{enumerate}
\item Pre-train the full language model from scratch on available large monolingual text corpora (e.g English) using language modeling training objective (for BERT it is masked language modeling);
\item Create a copy of the pre-trained model and replace the embeddings of the original with new embeddings for the target language;
\item Continue the pre-training of the modified original on target language monolingual corpora for model embeddings while freezing (not updating) all other layers using the same training objective;
\item Fine-tune the copy on the downstream task dataset while keeping the embeddings frozen;
\item Swap the embeddings of the fine-tuned copy with embeddings of the original model obtained after continued pre-training on the target language corpora.
\end{enumerate}
The major advantage of the described algorithm is that the continued pre-training step requires much less data than initial pre-training from scratch as it requires training only a fraction of model parameters which reduces model optimization task complexity and thus has faster convergence (Kaplan, 2020). The main hypothesis is that task-solving knowledge is language agnostic and it was confirmed in the original experiments (Artetxe, 2019) for natural language understanding and document classification tasks. However, the authors noted that fine-tuning on downstream tasks with frozen embeddings is not enough for proper embedding swap alignment and additional embedding transformations or special embedding utilization penalties are required to maximize the efficiency of target language vocabulary processing. As a possible solution to the embedding alignment problem Chen et al. (2023) proposed using a special pre-training regime with active embedding forgetting to force the language model to accumulate the knowledge in intermediate layers. The downside of such an approach is that we must have full control on the initial pre-training which is not possible for state-of-the-art LLMs obtained by training on high quality proprietary datasets with immense computational budget. 

We argue that embedding swap alignment can be achieved without special training procedures by leveraging the fine-tuning parameter update trajectory. Ilharco et al. (2023) showed that the fine-tuning trajectory may be approximated with linear transformations of base model parameters which can be derived from parameter decomposition of fine-tuned variants. Therefore, by finding appropriate linear transformations for embedding parameters we can approximate the results of a full language adaptation pipeline without involving the instruction-tuning dataset.

Formally, let $I, O$ be the input and output embeddings of LLM and $W$ a pseudo-linear approximation of intermediate LLM layer composition: 

\begin{equation}
LLM_{base}=I_{base}W_{base}O_{base} 
\end{equation}
    
Denote $D, U$ as linear embedding transformations that align original embeddings with the fine-tuned layers:  

\begin{equation}
LLM_{base\rightarrow inst}=I_{inst}W_{inst}O_{inst}=I_{base}D_{inst}W_{inst}U_{inst}O_{base}
\end{equation}
    
Since our target language embedding initialization strategy averages the embeddings of overlapping tokens in $I_{base}$ and $O_{base}$ we can formalize the initialization process with vocabulary transformation operation $T_{ru}$:

\begin{equation}
LLM_{base \rightarrow ru}=T_{ru}I_{base}W_{base}O_{base}T_{ru}^T=I_{ru}W_{base}O_{ru}
\end{equation}
    
Following the logic described above the fine-tune of language adapted base model $LLM_{ru\rightarrow inst}$. can be represented as the following:

\begin{equation}
LLM_{ru\rightarrow inst}=T_{ru}I_{base}D_{inst}^{ru}W_{ru\rightarrow inst}U_{inst}^{ru}O_{base}T_{ru}^{T}
\end{equation}
    
Now by assuming that the optimal $W_{ru\rightarrow inst}\approx W_{inst}$ we arrive at the final equation for propagation of continued pre-trained embeddings $I_{ru/cpt},O_{ru/cpt}$:

\begin{equation}
LLM_{ru/cpt\rightarrow inst}=I_{ru/cpt}D_{inst}^{ru}W_{inst}U_{inst}^{ru}O_{ru/cpt}
\end{equation}
    
The remaining variables $D_{inst}^{ru}, U_{inst}^{ru}$  are determined by chosen assumptions about embedding alignment properties. In our experiments we consider 3 options:
\begin{enumerate}
\item Direct embedding swap;
\item Overlapping token correction;
\item Vocabulary conversion projection.
\end{enumerate}

\subsubsection{Direct embedding swap}
Considering that most state-of-the-art LLMs are trained on multilingual datasets, it can be expected that their inner representations are tailored for language-agnostic text processing. Similarly to the original works on embedding-based knowledge transfer for encoder models we assume that the embedding layer carry only conceptual information i.e. we suppose $D_{inst}^{ru} =U_{inst}^{ru} = E$, where $E$ is an identity matrix.

\subsubsection{Overlapping token correction}
Since the considered LLMs are initially designed for multilingual text generation they have a basic set of the most common tokens for popular languages such as russian. The idea is to find the union $C = tokens^{old} \cap tokens^{new}$ of the original $tokens^{old}$ and language-adapted $tokens^{new}$ vocabularies and use this subset to reduce $I_X, O_X$ to the common components of embedding initialization $I_{X/com},O_{X/com}$ where $X \in \{base, inst\}$. This allows to approximate the embedding projections as $D_{inst}^{ru}\approx D_{inst}$ and $U_{inst}^{ru}\approx U_{inst}$:

\begin{equation}
D_{inst}^{ru}=I_{base/com}^{-1} I_{inst/com},
\end{equation}

\begin{equation}
U_{inst}^{ru}=O_{inst/com}O_{base/com}^{-1},
\end{equation}

\begin{equation}
I_{X/com}=[I_X^{idx(t)}]_{t\in C},
\end{equation}

\begin{equation}
O_{X/com}=[O_X^{idx(t)}]_{t\in C}.
\end{equation}

where $idx(t)$ is a function that maps token $t$ to its respective position in the embedding matrix. It must be noted that $I_{X/com},O_{X/com}$ matrices are likely to be not invertible and thus their inversion must be approximated with least squares problem solvers.

\subsubsection{Vocabulary conversion projection}
Since embedding initialization transformation $T_{ru}$ is universal for both base and fine-tuned models we can derive an alternative equation for obtaining language-adapted instruction-tuned LLM:

\begin{equation}
LLM_{inst\rightarrow ru}=T_{ru}I_{inst}W_{inst}O_{inst}T_{ru}^T
\end{equation}

By assuming that both variants of instruction-tune adaptation are equivalent $LLM_{ru\rightarrow inst}=LLM_{inst\rightarrow ru}$ we obtain the following formulae for embedding alignment:
\begin{equation}
D_{inst}^{ru}=(T_{ru}I_{base})^{-1}T_{ru}I_{inst}
\end{equation}

\begin{equation}
U_{inst}^{ru}=O_{inst}T_{ru}^{T}(O_{base}T_{ru}^{T})^{-1}
\end{equation}
    
Similarly to the previous alignment method the calculation of transformation matrices involves least square problem solvers for finding the pseudo-inversion of non-invertible matrices. This is the main reason why vocabulary transformation $T_{ru}$ should not be isolated. The pilot experiments showed that such simplification increases the error margin of alignment transformations which lowers the quality of embedding propagation procedure.

\subsection{Darumeru benchmark}
Existing LLM benchmarks for Russian language (Fenogenova, 2024) do not expose the testing data labels for local evaluation. On one hand such an initiative is reasonable amid the rising trend of training on test data which renders the LLM ranking results meaningless. On the other hand hidden test labels means that the evaluation requires having an online connection to the benchmark system which prevents evaluation in offline computational environments thus postponing the evaluation until the end of training session. Moreover lack of access to test labels makes it impossible to classify the type of prediction errors thus limiting the post-training quality analysis.

To address the issue we developed a new benchmark framework that focuses on quick and informative LLM text generation quality evaluation. This benchmark consists of combinations of open splits of datasets from MERA (Fenogenova, 2024), mmlu\_ru / mmlu\_en, RuCoLA (Mikhailov, 2022), as well as new datasets for text generation assessment - 17 datasets total. A more detailed description of each dataset is given in the following sections. 

\subsubsection{Framework}
The evaluation framework utilizes message format to ensure compatibility with both pre-trained and instruction-tuned LLMs. This means that all task data for the models is converted into a sequence of “user role”-”message content” pairs, from which the final prompt is constructed.  The framework supports tasks that require estimating the probability of the next token, generation, or logsoftmax for the entire generated sequence. The evaluation can be carried out directly in a conventional Transformers model training environment or via VLLM specialized model inference servers.

\subsubsection{DaruMERA and DaruMMLU}
We composed DaruMERA from the following MERA datasets: MultiQ, PARus, RCB, RWSD, USE, ruOpenBookQA, ruWorldTree. For better language understanding evaluation we also added validation split of RuCoLA dataset.
For DaruMMLU part we separated ruMMLU (MERA) and complemented it with MMLU datasets from the NLP-Core-Team repository\footnote{https://github.com/NLP-Core-Team/mmlu\_ru}.
There are several changes to the original datasets:
\begin{enumerate} 
\item MultiQ version was augmented with additional gold answers. The existing labels do not correspond in form to the questions, as they were extracted from the text without proper preprocessing. The augmentation process consisted of passing the question and reference answer pairs to LLaMa-3-70B-Instruct model to rephrase the answer in accordance with the question; 
\item The ruMMLU version differs from the similar one in NLP-Core-Team repository in that it has few-shot examples common to all queries, regardless of the domain, and also uses not one fixed template, but several options as instructions; 
\item When calculating PARus, for each example the same example was generated, but with a different order of options, and only the case when the model predicts the correct option for both the direct and reverse order was considered a success.
\end{enumerate}
To measure the performance on PARus, RWSD, MMLU datasets we used accuracy metric. For RCB, ruOpenBookQA and ruWorldTree we averaged accuracy and F1-macro. For RuCoLa we used average of accuracy and Matthews Correlation Coefficient (MCC). For MultiQ we used the average of F1 and exact match metrics. For USE the normalized total grade was used. 

\subsubsection{DaruSum}
Most of the evaluation tasks aim to measure the model's text comprehension capabilities and global contextual knowledge which is required for proper prompt processing. However for text generation the model must be also capable of filtering the input text for the query relevant content to ensure that the user would receive the desired answer regardless of input format or size. Text summarization is the perfect evaluation task for such a case as it requires both filtering the input content and composing the answer from the salient fragments. 

There are two summarization settings: extractive and abstractive. Extractive summarization is a task of sentence saliency ranking where the summary is obtained by taking top-k ranked sentences. Abstractive summarization on the other hand is a text generation task where saliency ranking is integrated in the token sampling process as the model guides itself toward the most concise summary. While the abstractive setting has the higher preference it is hard to distinguish automatically the suboptimal content filtering from the text generation errors. At the same time constraining the text generation process to input fragments such as sentences basically reduces the task to extractive summarization. Thus to evaluate content filtering accuracy and text generation quality it is sufficient to evaluate the abstractive summarization in free and constrained generation settings.

For the summarization dataset we chose Gazeta (Gusev, 2020) which has established itself as the standard for Russian automatic summarization evaluation. To improve the accuracy of evaluation procedure we derived an example filtering protocol that all reference summary content can be inferred from the input document. Since LLaMa-3-70B showed high human agreement in LLM evaluation\footnote{https://github.com/tatsu-lab/alpaca\_eval} we employed it as the example correctness evaluator and tasked it to find all citations that support the summary sentence. We filtered out all examples that had more than 20\% of unsupported summary sentences and mapped found citations to document sentences, thus producing accurate extractive labels. To adapt the task for a few-shot setting which is limited by context window limitations we compressed the documents by dropping the paragraphs that had no extractive summary labels. To account for LLM text generation length variance (Dubois, 2024) as the metric for abstractive and extractive settings we chose average of ROUGE-1 and ROUGE-2 recall and R-precision respectively.

\subsubsection{DaruCopy}
When replacing the LLM vocabulary it is important that it learns to fully utilize new tokens.  The input token embeddings are responsible for conveying the text meaning which can be evaluated by natural language understanding tasks such as MMLU. In contrast, the output token embeddings are used to find the closest semantic meaning to the current neural network state which depends on contextual history. As a consequence, in creative tasks this state is unstable and LLM tends to generate rarer tokens. At the same time, in the tasks where the LLM is required to reuse the input context the network state is expected to fall into semantic clusters of tokens that are present in the input sequence. Following that logic by prompting the LLM to produce a copy of the input text we can evaluate its token generation efficiency. 

We used Wikipedia articles of different genres to collect copy task datasets for English and Russian languages involving 2 copy settings: sentence-wise and paragraph-wise. The former setting assesses the LLM alignment with tokenization algorithm which is calculated as the ratio of the length of the original text to the generated text in tokens. In paragraph setting we evaluate the overall text generation stability by measuring the percentage of generations in which the ratio of longest common subsequence (lcs) tokens to all paragraph tokens is greater than 99\% (1\% is left for spacing errors). Deviation from 99\% amid the high sentence copy scores indicates that the model tends to confuse tokens and thus can hallucinate context in creative tasks which is the major reliability concern for practical applications.

\subsubsection{Benchmark parameters}
When calculating the benchmark metrics, the following parameters were set: batch size 8, sequence length 4096, 5-shot for foundation models and zero-shot for instruct models.

\subsection{Experiment setting}
We conducted adaptation experiments with two models: Mistral-7B-v0.1 (Jiang, 2023) and LLaMa-3-8B (Dubey, 2024).

\subsubsection{Continued Pre-training}
Training dataset for tokenization and continued pre-training consists of documents from the following domains: Russian Wikipedia, English Wikipedia, Habrahabr, Pikabu, Fiction, News, Educational literature.
The documents were deduplicated using Locality Sensitive Hashing Minhash algorithm. We removed metadata, links, comment sections and badly formatted documents to improve vocabulary distribution and reduce the number of grammatically incorrect examples. To reduce the semantic noise we restricted the vocabulary to Cyrillic and Latin languages and stripped non-standard symbols like emoji or logograms (e.g. Chinese characters) using UTF-8 normalization.
For training, texts were sampled with increased weights for Wikipedia, educational and scientific literature. Additionally, to feed texts into the language model, we ensured that each sample began either with a new document or with a new paragraph.

\textbf{Tokenization parameters.} We trained BPE and Unigram tokenizers with 32000 and 128000 tokens for Mistral-7B and LLaMa-3-8B respectively. For Extended tokenizer, we extended the original tokenizers to 55328 and 174816 tokens using new Russian-adapted BPE vocabularies for corresponding models. Since LLaMa-3-8B tokenization vocabulary is likely to be extensive we created an Optimized version, where we shrunk the original BPE vocabulary to 64000 tokens and then merged with top 64000 most common tokens from new BPE vocabulary, resulting in 114504 tokens.

\textbf{Hyperparameters.} During continued pre-training we used the following hyperparameters: Total Batch Size: 256; Block Size: 1024; Weight Decay: 0.1; Scheduler: Cosine; Warmup Steps: 100; Epochs: 1.

We tested 4 different learning rates: 2e-5, 5e-5, 1e-4, 2e-4 for each model and tokenization on 20\% of all continued pre-training dataset. Based on benchmark results, we chose a learning rate equal to 1e-4 for all Mistral-7B models, and learning rate equal to 2e-4 for LLaMa-3-8B models. It is important to note that the efficiency of model adaptation showed a significant dependence on the learning rate, especially for LLaMa-3-8B based models. 

\subsubsection{Case Study: Self-Calibration}
For the cases of full vocabulary substitution where the model learns to rewire all new embeddings virtually from scratch the propagation process may have lower efficiency as the difference between instruct-tuned and language-adapted embeddings may be dramatic. The logical solution is to synthesize self-instruct data using the original instruct-tuned LLM and then use it to calibrate the language-adapted version. To generate the examples, we used prompts from Saiga instruction dataset and used greedy decoding to get the most likely answer from instruct-tuned LLM viewpoint. Then we asked LLaMa-3-70B to evaluate the quality of synthesized pairs in terms of grammar and relevance on a 5-point grading scale. All examples that received a score less than 4 were discarded which left us 13531 calibration examples. 

Since calibration examples are native for LLM inner semantic representations there is a risk that instead of alignment the model may revert back to the original tokenization behavior which prioritizes smaller but more familiar tokenization chunks. To avert such a scenario we leverage the fact that all modern LLMs are pre-trained on Wikipedia articles in such a manner that their embedding representations are aligned with Wikipedia concepts. By asking the fine-tuned model to repeat a Wikipedia article token by token we force the model to recall its pre-training memory and thus to propagate the activation signals respective to the concepts in the article to embeddings of optimal tokens of new tokenization. Following that logic we supplemented the self-instruct dataset with 10000 article-copy task examples, obtained from the part of Wikipedia that has no overlap with our pre-training or benchmark datasets.

We found the following LoRA-tuning settings to be optimal for calibration procedure: Rank: 8; Alpha: 1; Learning Rate: 2.5e-5; Weight Decay: 0.1; LoRa target modules: first and last transformer layers; LoRa modules to save: lm\_head, embed\_tokens; Max Sequence Length: 8096 (i.e. max context length); Total Batch Size: 64; Epochs: 1.

\subsubsection{Case Study: Continued Instruction-Tuning Calibration}
In addition to the self-calibration experiments, we decided to test how continued instruction-tuning on the high-quality Russian instruction dataset would affect the final performance. For this experiment we choose Saiga dataset\footnote{https://huggingface.co/datasets/IlyaGusev/saiga\_scored} which is considered to be the best open-source option for Russian language. We also investigated the impact of adding a small number (2000) of special instructions to the dataset, the purpose of which is to copy a large text from Wikipedia
To fine-tune the models we used LoRA adapters with Saiga-recommended hyperparameter settings which is the following: Rank: 32; Alpha: 16; Learning Rate: 5e-5; Weight Decay: 0.05; LoRa target modules: attention, mlp; LoRa modules to save: lm\_head; Max Sequence Length: 4096; Total Batch Size: 128; Epochs: 1.

\section{Results}
\subsection{Open-source LLM Benchmark}
To establish a baseline we benchmarked popular instruct-tuned LLMs (see Table 1): Openchat 3.5, LLaMa-3 (instruct) (Dubey, 2024), Saiga (Gusev, 2023), Vikhr (Nikolich, 2024), Qwen-2, Mistral Nemo (Jiang, 2023). As expected the largest model, Mistral Nemo, has the highest zero-shot performance. Smaller counterparts have the same score margin. However, Qwen-2 7B manages to outperform Mistral Nemo in MMLU tasks while falling behind on text generation robustness tests of DaruSum and DaruCopy. Vikhr-5.2 similarly has the same score on DaruMERA as Mistral Nemo. Considering the LLM scaling laws (Kaplan, 2020) and the performance gap with state-of-the-art sub-10B parameter LLM, LLaMa-3, this observations suggest that some parts of MMLU and MERA datasets were leaked to training data of Vikhr-5.2 and Qwen-2 7B. 

\begin{table}[h!]
\centering
\caption{Darumeru zero-shot evaluation results for popular open-source instruct-tuned models.}
\label{tab:table1}
\resizebox{1.0\textwidth}{!}{
\begin{tabular}{lcccccc}
\toprule
Model & Micro-Avg & DaruMMLU & DaruMERA & DaruSum & DaruCopy (EN) & DaruCopy (RU) \\
\midrule
Openchat 3.5 (Mistral-7B) & 0,607 & 0,543 & 0,526 & 0,322 & 0,999 & 0,917 \\
\cmidrule(r){2-7}
LLaMa-3-8B (Instruct) & 0,610 & 0,571 & 0,510 & 0,322 & \textbf{1,000} & 0,972 \\
\cmidrule(r){2-7}
Saiga (LLaMa-3-8B) & 0,608 & 0,574 & 0,514 & 0,320 & 0,995 & 0,939 \\
\cmidrule(r){2-7}
Vikhr-5.2 (Mistral-7B) & 0,587 & 0,494 & \textbf{0,573} & 0,308 & 0,959 & 0,693 \\
\cmidrule(r){2-7}
Qwen-2 7B & 0,613 & \textbf{0,624} & 0,548 & 0,300 & 0,938 & 0,842 \\
\cmidrule(r){2-7}
Mistral Nemo (12B) & \textbf{0,639} & 0,592 & \textbf{0,576} & 0,320 & 0,998 & 0,924 \\
\midrule
\multicolumn{7}{c}{Ours} \\
\midrule
\makecell[l]{Openchat 3.5 + \\LEP-Extended + \\calibration (best)} & $0,632^{\uparrow}$ & 0,541 & $0,563^{\uparrow}$ & 0,321 & \textbf{1,000} & \textbf{$0,989^{\uparrow}$} \\
\cmidrule(r){2-7}
\makecell[l]{LLaMa-3-8B (Instruct) + \\LEP-Extended + \\calibration (best)} & $0,618^{\uparrow}$ & $0,565^{\downarrow}$ & $0,521^{\uparrow}$ & \textbf{$0,339^{\uparrow}$} & \textbf{1,000} & \textbf{$0,984^{\uparrow}$} \\
\bottomrule
\end{tabular}}
\end{table}

\subsection{Vocabulary Adaptation and Continued Pre-Training}
Following our initial benchmark results we focused on Russian adaptation of the foundation models of the most performant instruct-tunes: Mistral-7B and LLaMa-3-8B. To evaluate the language-adaption results we used few-shot in-context-learning as the models are not used to interpreting the instructions directly.

Figure 2 shows the Darumeru score dynamic throughout the continued pre-training process. In case of Mistral-7B the vocabulary substitution methods such as BPE and Unigram almost exhaust the training examples converging to the optimum at the final 10k training steps. In contrast LLaMa-3-8B is more robust to vocabulary adaptation methods as they all tend to converge in the middle of a training session at 20-30k steps. Since the full dataset size is 96 GB we can conclude that 40 GB of texts is the minimum required for the good performance of Russian adapted embeddings.

\begin{figure}
  \centering
  \includegraphics[scale=0.4]{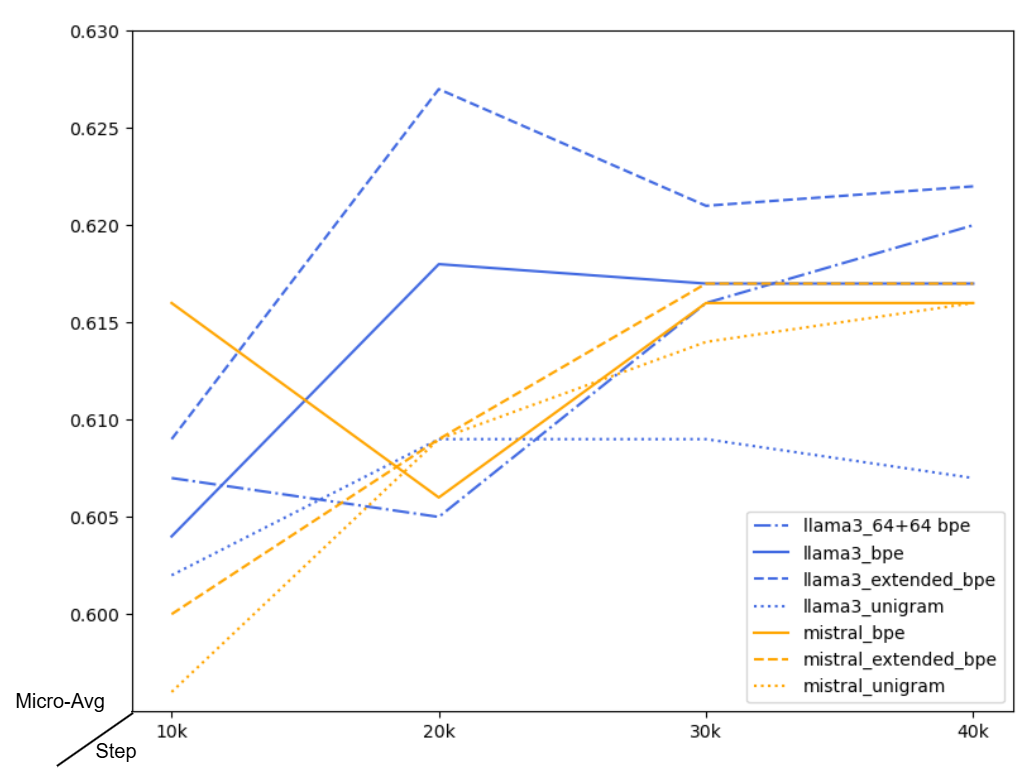}
  \caption{Micro average benchmark score dynamic throughout training}
  \label{fig:fig2}
\end{figure}

In Table 2 we report the detailed results of the best performing checkpoints. As expected, vocabulary extension methods such as Extended and Optimized have the lowest optimization difficulty as they show the highest language-adaptation scores. For Mistral-7B all language adaptations significantly outperform the original foundation, however the difference between their tokenization efficiency (symbols per token) and average task-performance may be considered marginal. For LLaMa-3-8B only Extended variants managed to reach the original LLM benchmark scores mainly falling behind on DaruMMLU tasks. Most tokenization-efficient variants, BPE and Unigram, considerably lag behind, losing in DaruMERA and DaruSum. We assume that vocabulary substitution in case of BPE and Unigram has a major impact on language understanding and that in their case continued pre-training of embeddings-only is not sufficient for proper semantic alignment and additional tuning procedures are required.

\begin{table}[h!]
\centering
\caption{Darumeru few-shot evaluation results for best language-adaptation checkpoints.}
\label{tab:table2}
\resizebox{1.0\textwidth}{!}{
\begin{tabular}{lcccccccc}
\toprule
Model & Vocab & \multicolumn{1}{c}{Symbols per token} & \multicolumn{1}{c}{Micro-Avg} & DaruMMLU & DaruMERA & DaruSum & DaruCopy (EN) & DaruCopy (RU) \\
\midrule
\multirow{4}{*}{Mistral-7B} &
Original & 2,44 & 0,604 & \textbf{0,545} & 0,504 & 0,307 & \textbf{1,000} & \textbf{1,000} \\
& BPE & \textbf{3,76} & 0,616 & 0,528 & 0,537 & \textbf{0,316} & 0,995 & 0,984 \\
& Unigram & \textbf{3,78} & 0,614 & 0,516 & \textbf{0,544} & 0,311 & 0,995 & 0,960 \\
& Extended & \textbf{3,77} & \textbf{0,617} & 0,538 & 0,532 & \textbf{0,314} & \textbf{1,000} & 0,995 \\
\cmidrule(r){2-9}
\multirow{5}{*}{LLaMa-3-8B} &
Original & 2,89 & \textbf{0,629} & \textbf{0,582} & \textbf{0,547} & \textbf{0,326} & 0,980 & 0,982 \\
& BPE & \textbf{4,40} & 0,618 & 0,561 & 0,532 & 0,321 & \textbf{1,000} & 0,963 \\
& Unigram & \textbf{4,35} & 0,609 & 0,560 & 0,517 & 0,316 & \textbf{1,000} & 0,951 \\
& Extended & 3,78 & \textbf{0,627} & 0,560 & \textbf{0,550} & \textbf{0,325} & 0,980 & 0,983 \\
& Optimized & 3,40 & 0,620 & 0,552 & 0,536 & 0,323 & 0,981 & \textbf{0,989} \\
\bottomrule
\end{tabular}}
\end{table}

\subsection{Learned Embedding Propagation}
The results of complete Learned Embedding Propagation (LEP) are reported in Table 3. For each adapted vocabulary construction option (BPE, Unigram, Extended and Optimized) we test 3 methods: Direct Embedding Swap (\textbf{Swap}), Overlapping Token Correction (\textbf{Overlap}) and Vocabulary Conversion (\textbf{Conversion}). For embedding donor model we used best continued pre-training checkpoints (see Table 2).
\begin{table}[h!]
\centering
\caption{Darumeru zero-shot evaluation results for Learned Embedding Propagation methods.}
\label{tab:table3}
\resizebox{1.0\textwidth}{!}{
\begin{tabular}{lccccccc}
\toprule
\textbf{Vocab} & \textbf{LEP method} & \textbf{Micro-Avg} & \textbf{DaruMMLU} & \textbf{DaruMERA} & \textbf{DaruSum} & \textbf{DaruCopy (En)} & \textbf{DaruCopy (Ru)} \\
\midrule
\multicolumn{8}{c}{OpenChat-3.5} \\
\midrule
BPE & Swap & \textbf{0,587} & \textbf{0,528} & \textbf{0,526} & 0,277 & 0,988 & \textbf{0,829} \\
BPE & Overlap & 0,584 & 0,525 & 0,523 & 0,281 & 0,986 & 0,818 \\
BPE & Conversion & 0,583 & 0,526 & 0,524 & \textbf{0,284} & \textbf{0,993} & 0,791 \\
\cmidrule(r){2-8}
Unigram & Swap & 0,556 & \textbf{0,517} & 0,517 & 0,282 & 0,985 & 0,614 \\
Unigram & Overlap & \textbf{0,572} & 0,514 & \textbf{0,534} & 0,297 & 0,981 & \textbf{0,680} \\
Unigram & Conversion & 0,565 & 0,515 & 0,519 & \textbf{0,301} & \textbf{0,999} & 0,651 \\
\cmidrule(r){2-8}
Extended & Swap & \textbf{0,608} & \textbf{0,535} & \textbf{0,540} & 0,298 & \textbf{0,999} & \textbf{0,907} \\
Extended & Overlap & \textbf{0,607} & \textbf{0,535} & \textbf{0,539} & \textbf{0,307} & \textbf{0,999} & 0,898 \\
Extended & Conversion & \textbf{0,609} & \textbf{0,535} & \textbf{0,541} & \textbf{0,306} & \textbf{0,999} & \textbf{0,909} \\
\midrule
\multicolumn{8}{c}{LLaMa-3-8B (instruct)} \\
\midrule
BPE & Swap & 0,565 & \textbf{0,544} & 0,486 & \textbf{0,317} & \textbf{0,999} & 0,729 \\
BPE & Overlap & \textbf{0,569} & \textbf{0,546} & \textbf{0,489} & 0,314 & \textbf{0,999} & \textbf{0,753} \\
BPE & Conversion & \textbf{0,570} & \textbf{0,546} & \textbf{0,490} & \textbf{0,318} & \textbf{0,999} & \textbf{0,754} \\
\cmidrule(r){2-8}
Unigram & Swap & \textbf{0,582} & \textbf{0,545} & \textbf{0,488} & \textbf{0,313} & \textbf{0,999} & 0,865 \\
Unigram & Overlap & 0,580 & \textbf{0,545} & 0,482 & \textbf{0,314} & \textbf{0,999} & 0,876 \\
Unigram & Conversion & \textbf{0,584} & \textbf{0,545} & \textbf{0,488} & \textbf{0,315} & 0,994 & \textbf{0,889} \\
\cmidrule(r){2-8}
Extended & Swap & 0,592 & \textbf{0,557} & 0,498 & \textbf{0,319} & 0,969 & 0,921 \\
Extended & Overlap & \textbf{0,597} & \textbf{0,556} & \textbf{0,504} & \textbf{0,321} & 0,964 & \textbf{0,936} \\
Extended & Conversion & \textbf{0,597} & \textbf{0,556} & 0,501 & 0,318 & \textbf{0,994} & 0,921 \\
\cmidrule(r){2-8}
Optimized & Swap & 0,594 & \textbf{0,554} & \textbf{0,499} & \textbf{0,327} & 0,970 & \textbf{0,928} \\
Optimized & Overlap & 0,586 & \textbf{0,553} & 0,495 & 0,323 & 0,925 & 0,925 \\
Optimized & Conversion & \textbf{0,598} & \textbf{0,555} & \textbf{0,500} & 0,324 & \textbf{0,995} & \textbf{0,928} \\
\bottomrule
\end{tabular}}
\end{table}

For Mistral-7B and OpenChat 3.5 the embedding propagation results have large variance depending on the chosen tokenization algorithm for Russian vocabulary. In case of BPE, which is the same algorithm used for the original, the trained embedding for new vocabulary has the highest alignment with instruct-tuned counterpart in case of direct embedding swap. In case of more morphologically correct Russian tokenization, Unigram, overlap projection has the highest average task performance. However, if we look at group-wise scores it becomes evident that conversion is a better option as it leads in every task but DaruCopy (Ru) where all unigram conversion variants are experiencing issues. The conventional vocabulary extension also leans towards conversion projection and has the best overall task performance among all vocabularies even outperforming the original OpenChat 3.5.

For LLaMa-3-8B embedding conversion is more straightforward. For all tokenization variants the conversion projection yields the best results, however, unlike the Mistral-7B LEP none of embedding propagations manage to reach the original LLaMa-3-8B (instruct) quality. The significant performance degradation is observed among all task groups with DaruCopy taking the biggest hit. Moreover, despite being the original tokenization algorithm, BPE-build Russian vocabulary has the lowest embedding compatibility with instruction-tune having the largest score gap. While the vocabulary Optimized variant has lower vocabulary size limit it maintains the same quality level as Extended and comparing the conversion projections the former has better DaruSum and DaruCopy solving capabilities.

There are several implications of the observations. First is that the Vocabulary Conversion LEP algorithm is likely to be the most efficient solution for the majority of embedding projection scenarios in some cases even being sufficient for recovering the original instruction-tuned performance. Secondly, while Unigram tokenization vocabulary may be considered morphologically correct for Russian language it is inferior to Extended and Optimization options as it requires full vocabulary substitution, which, considering unstable BPE performance, creates the largest disparity between embedding and inner-layer semantic representation. The tokens removed in the Optimized variant seem to be unimportant for Russian task-solving capabilities as it manages to outperform Extended tokenization which completely retains the original vocabulary. The performance gap in LEP LLaMa-3-8B (instruct) is likely to be the consequence of proprietary instruction-tuning dataset which was large enough to align the embedding semantics with instruction-following tasks (Dubey, 2024). Another hypothesis is that the original LLM underwent human preference alignment procedure which aims to block text generation of harmful answers at the cost of necessary reasoning limitations and as a consequence has a habit of blocking potential malicious semantics originating from input embeddings which in turn inhibits text comprehension capabilities. 

\subsection{Case Study: Self-Calibration}
In self-calibration experiments we focused on closing the gap of best LEP LLaMa-3-8B instruct models (Table 4, self-calibration). As expected the performance of DaruCopy tasks improved substantially, practically reaching the perfect reliability levels. DaruSum also saw the improvements as the improved citation capabilities are beneficial for composing concise summaries. Other tasks however did not improve much and in the case of the weakest vocabulary adaptation, Unigram, saw a significant decline in benchmark scores. 

We suspect that the self-calibration data promotes closed-mind reasoning as training on the most probable answers biases the model towards generic vocabulary which had the highest frequency in the training data. As a consequence the comprehension of rarer domain-specific concepts which are present in MMLU and MERA datasets may be inhibited due to increased tendency of using more common language. The issue can be alleviated by more complex example sampling procedures such a beam search or multi-candidate generation with post-generation ranking with larger state-of-the-art LLMs such as GPT-4 or LLaMa-3-405B. 

\subsection{Case Study: Continued Instruction-Tuning Calibration}
Our experiments on continued instruction-tuning calibration approach, presented in Table 4, showed that the additionally fine-tuned LEP adapted models achieve and in some cases outperform the original models. Adding 2000 instructions for copying long texts to the instructional dataset has a positive effect in almost all cases. Moreover, the obtained models are more effective when used in the Russian language, and the loss of initial knowledge in the case of our method is minimal, compared to conventional instruct-tuning.

\begin{table}[h!]
\centering
\caption{Benchmark results for model calibration schemes of Conversion LEP models}
\label{tab:table4}
\resizebox{1.0\textwidth}{!}{
\begin{tabular}{lccccccc}
\toprule
Model & Fine-tuning data & Micro-Avg & DaruMMLU & DaruMERA & DaruSum & DaruCopy (EN) & DaruCopy (RU) \\
\midrule
\multicolumn{8}{c}{OpenChat-3.5} \\
\midrule
\multirow{3}{*}{Original model} & - & 0,607 & \textbf{0,543} & 0,526 & 0,322 & \textbf{0,999} & 0,917 \\
& saiga d7 & 0,611 & 0,540 & \textbf{0,528} & \textbf{0,325} & \textbf{0,999} & 0,945 \\
& +copy task & \textbf{0,615} & 0,541 & 0,524 & 0,324 & \textbf{1,000} & \textbf{0,995} \\
\cmidrule(r){2-8}
\multirow{3}{*}{Unigram} & - & 0,565 & 0,515 & 0,519 & 0,301 & \textbf{0,999} & 0,651 \\
& saiga d7 & 0,599 & \textbf{0,532} & 0,556 & 0,316 & \textbf{0,999} & 0,754 \\
& +copy task & \textbf{0,630} & 0,530 & \textbf{0,559} & \textbf{0,321} & \textbf{1,000} & \textbf{0,999} \\
\cmidrule(r){2-8}
\multirow{3}{*}{Extended} & - & 0,609 & 0,535 & 0,541 & 0,306 & \textbf{0,999} & 0,909 \\
& saiga d7 & 0,616 & \textbf{0,543} & \textbf{0,566} & 0,319 & \textbf{0,999} & 0,845 \\
& +copy task & \textbf{0,632} & 0,541 & 0,563 & \textbf{0,321} & \textbf{1,000} & \textbf{0,989} \\
\midrule
\multicolumn{8}{c}{LLaMa-3-8B (instruct)} \\
\midrule
\multirow{3}{*}{Original model} & - & 0,610 & 0,571 & 0,510 & 0,322 & \textbf{1,000} & 0,972 \\
& saiga d7 & 0,615 & \textbf{0,576} & 0,512 & 0,329 & \textbf{1,000} & 0,983 \\
& +copy task & \textbf{0,616} & 0,575 & \textbf{0,513} & \textbf{0,332} & \textbf{1,000} & \textbf{0,995} \\
\cmidrule(r){2-8}
\multirow{3}{*}{Extended} & - & 0,597 & 0,556 & 0,501 & 0,318 & 0,994 & 0,921 \\
& self-calibration & 0,606 & 0,552 & 0,512 & 0,321 & \textbf{1,000} & 0,958 \\
& saiga d7 & 0,614 & \textbf{0,568} & 0,519 & 0,338 & 0,995 & 0,961 \\
& +copy task & \textbf{0,618} & 0,565 & \textbf{0,521} & \textbf{0,339} & \textbf{1,000} & \textbf{0,984} \\
\cmidrule(r){2-8}
\multirow{3}{*}{Optimized} & - & 0,598 & \textbf{0,555} & 0,500 & 0,324 & 0,995 & 0,928 \\
& self-calibration & 0,601 & 0,550 & 0,501 & 0,325 & \textbf{1,000} & 0,950 \\
& saiga d7 & 0,611 & \textbf{0,555} & 0,515 & 0,336 & \textbf{1,000} & 0,971 \\
& +copy task & \textbf{0,617} & \textbf{0,555} & \textbf{0,522} & \textbf{0,339} & \textbf{1,000} & \textbf{0,989} \\
\bottomrule
\end{tabular}}
\end{table}

\subsection{Examples}
We also investigated how the models' responses changed depending on the stage: original model, LEP, LEP + calibration (Figure 3).
\begin{figure}
  \centering
  \includegraphics[scale=0.2]{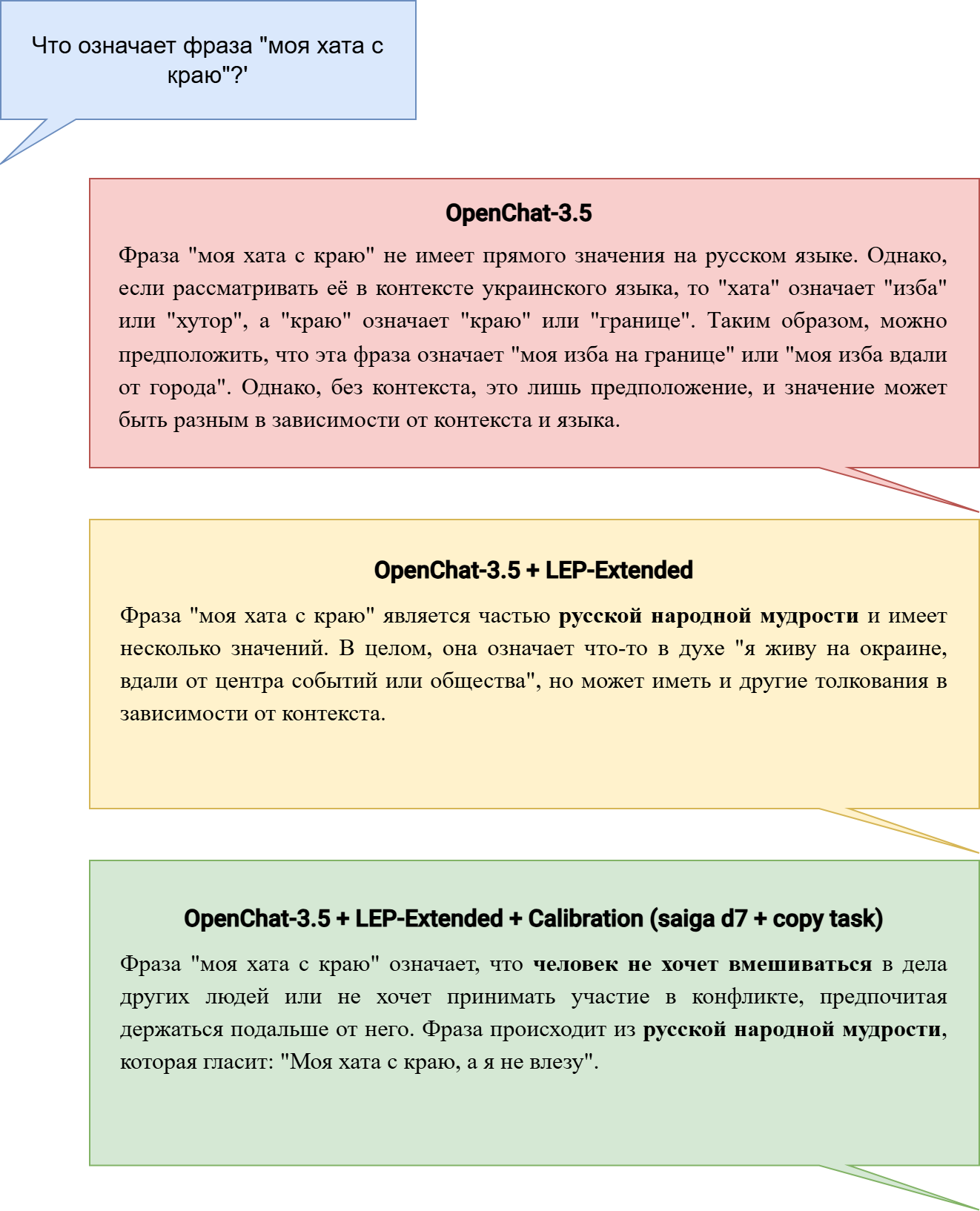}
  \caption{An example of generation using the OpenChat-3.5 model and its adapted versions.}
  \label{fig:fig3}
\end{figure}

From the example, it can be seen that the original model did not correctly perceive the question at all. The LEP model already answers more correctly, but does not take into account that this is a phraseological unit. The calibrated model already answers the question most correctly among the three versions of the model, paying attention to the true meaning of the phrase.

\section{Discussion}
\subsection{LLM benchmark results for Russian language}
Results presented in Table 1 demonstrate that fine-tuning of open-source state-of-the-art LLMs on Russian focused instruction datasets commonly leads to performance drops in language understanding. This phenomenon was initially observed within Ru-Arena-General\footnote{https://huggingface.co/spaces/Vikhrmodels/arenahardlb} and Chatbot Arena\footnote{https://lmarena.ai/} benchmarks, however, due to their open-question format it was hard to separate generation errors from bad user prompting. Closed-question benchmarks such as MERA (Fenogenova, 2024), which was used as the basis of Darumeru, can not reliably detect language processing degradation due to the possibility of benchmark hacking. Benchmark hacking is a procedure of fine-tuning on benchmark solutions or similar data which is viewed as a variant of cheating in the context of LLM benchmarks. Usually developers of LLM models do not intend to resort to such poor practice and on the contrary make an additional effort to remove any possible benchmark data from the overall LLM training data pool. At the same time detecting benchmark related data-leaks is a labor-intensive task as it requires checking training data not just for exact matches but also for any possible paraphrases which includes translating examples to other languages. 

Our Darumeru benchmark addresses the limitation of closed-question format with newly introduced tasks for text summarization (DaruSum) and tokenization diagnostic (DaruCopy). DaruSum requires two crucial task-solving elements, proper text analysis and good text writing skills. Any performance drops in this benchmark subset indicate problems with text understanding or text generation. DaruCopy distinguishes between the two by exclusively evaluating the latter by reducing the task to explicitly broadcasting the original context without any analysis or paraphrasing. Consequently, lower DaruCopy scores indicate a reasoning conflict within the LLM logic as the model fails to follow simplest task directive of text copying. These two subsets of Darumeru benchmark show that LLaMa-3-8B is a more reliable choice for Russian processing tasks than Saiga or Vikhr-5.2 despite their Russian language specialization which contrasts with the results of MERA benchmark (Fenogenova, 2024). While MERA results of Saiga lie within standard deviation the results of Vikhr-5.2 clearly suggest the case of benchmark hacking.

\subsection{Language adaptation strategy}
During development of our LLM Russian adaptation pipeline we made several design choices which were explored in previous works. First of all, we assumed that tokenization knowledge and the ability to use new tokens is stored in input embeddings and LM head layers of LLM. Several works (Cui, 2023; Tikhomirov, 2023; Nikolich 2024; Nguyen, 2024) demonstrated that language-adaptation of these subset of layers only is insufficient for proper language understanding and thus subsequent instruction-tuning of such models leads to suboptimal results. At the same time it was shown (Tikhomirov, 2024) that there is no significant difference between language-adaptation of all-layers and dual-stage approach, when embedding and LM-head training process is complemented with subsequent training of other layers. Results reported in Table 2 reinforce this claim as the first stage of dual-stage approach proves to be efficient enough to substantially improve Russian language comprehension of Mistral-7B model. However, LLaMa-3-8B post-adaptation scores suggest that the necessity of inner-layer training is dictated by the original LLM Russian linguistic skills which are effectively captured by the DaruMERA subset of our benchmark. Learned Embedding Propagation procedure results (see Table 3) also reflect this observation as Mistral-7B showed highest language-knowledge transfer efficiency. 

Whether layer discrepancy can be alleviated by instruction-tuning we explored in our calibration experiments. Instruction-tuning on target language often improves token utilization and boosts language comprehension (Gusev, 2023; Wei, 2023; Nikolich, 2024). We see a similar trend in Table 4. By training the original non-adapted instruction-tuned versions of LLMs on Saiga dataset (Gusev, 2023) we enhanced Russian task-solving capabilities which boosted benchmark scores. Applying the same procedure to our LEP models (saiga d7) we retain the positive effect at increased rates with the scores higher than of the original models which were the subjects of LEP knowledge transfer. The drawback of instruction-tuning on Russian instruction datasets is that we inevitably disturb the original knowledge that was gained in prior training (Tejaswi, 2024). We attempted to address the issue by training on the answers generated by the original LLM (self-calibration) rather than using the original references from the Saiga dataset. However for LLaMa-3-8B instruct we did not see noticeable improvement in any LLM capabilities besides tokenization utilization (DaruCopy). This result is likely due to lack of generation quality of our self-calibration synthesized examples which during our manual inspection revealed to carry much simpler Russian language logic and vocabulary. Considering that Saiga is a prime example of GPT-4 reference synthesis (Taori, 2024) we hypothesize that by utilizing more advanced sampling techniques and better example quality evaluation protocols we may collect a reference dataset with the similar features without employment of other datasets or third-party models.

\section{Limitations}
Despite the broad applicability of our method, this study has several limitations. First, the method requires that not only instructional versions of LLM but also their foundational versions be available, which is not always the case. Secondly, in the case of languages using hieroglyphs, initialization after tokenizer replacement can be quite weak due to lack of shared tokens and it is not known how much adaptation of embeddings can help with this. Another important point is that the focus of the knowledge transfer procedure was on preserving the original knowledge of the target model which is why the possible volume of transferred knowledge may be insufficient. However, since the methodology effectively adapts the model to the language, it is always possible to conduct an additional stage of continuous pretraining to acquire new knowledge.

\section{Conclusion}
In this paper, we proposed Learned Embedding Propagation (LEP), an improved approach to large language model (LLM) language adaptation that has minimal impact on LLM inherent knowledge while enabling transferring the language-adaptation knowledge directly to any instruct-tuned version, including the proprietary. Focussing on cost-efficiency of our method we derived 3 ad-hoc approaches for the embedding propagation: Direct Embedding Swap, Overlapping Token Correction and Vocabulary Conversion. To facilitate the development process of optimal Russian adaptation we introduced Darumeru, a train-time benchmark which focuses on text generation reliability. By analyzing the benchmark performance of popular instruction-tune LLMs and 4 vocabulary adaptation options we derived a recipe for the most cost-efficient procedure. Using the recipe and the proposed LEP methods we built language-adapted variants of sub-9B parameter state-of-the-art instruction-tuned LLMs, Openchat-3.5 and LLaMa-3-8B (Instruct). The evaluation results demonstrated that the Vocabulary Conversion LEP variants reproduce the performance levels of the original instruction-tuned LLM and in the case of OpenChat–3.5 even outperform while having all benefits of improved computational efficiency. To close the remaining gaps in task-solving performance we conducted case-study experiments on self-calibration and continued instruct-tuning alignment approaches which concluded with further language comprehension improvements and new benchmark records. The obtained results open new prospects for LLM language adaptation enabling cost-efficient utilization of any instruction-tuned models regardless of openness of their fine-tuning data with all the merits of the original version.

All our models\footnote{https://huggingface.co/RefalMachine}, code\footnote{https://github.com/RefalMachine/ruadapt}, benchmark and framework\footnote{https://github.com/RefalMachine/llmtf\_open} are open source and available under the original model licenses.

\section*{Acknowledgments}
The work of Mikhail Tikhomirov was supported by Noncommercial Foundation for Support of Science and Education ”INTELLECT”. The work of Daniil Chernyshev was supported by Noncommercial Foundation for Support of Science and Education ”INTELLECT”.  The research was carried out using the MSU-270 supercomputer of Lomonosov Moscow State University. \nocite{*}
\bibliographystyle{unsrt}  

\end{document}